# Deep Learning for Video-Based Assessment of Endotracheal Intubation Skills

Jean-Paul Ainam[a], Erim Yanik[b], Rahul Rahul[a], Taylor Kunkes[c], Lora Cavuoto[c], Brian Clemency[d], Kaori Tanaka[d], Matthew Hackett[e], Jack Norfleet[e], Suvranu De[b1]

[a]*Center for Modeling, Simulation, & Imaging in Medicine, Rensselaer Polytechnic Institute, NY, USA*
[b]*Florida Agriculture & Mechanical University – Florida State University College of Engineering, Tallahassee, FL, 32310, USA*
[c]*Department of Industrial and Systems Engineering, University at Buffalo, Buffalo, NY*
[d]*Department of Emergency Medicine, University at Buffalo, Buffalo, NY*
[e]*U.S. Army Futures Command, Combat Capabilities Development Command Soldier Center STTC, Orlando FL 32826, USA*

**Abstract**

Endotracheal intubation (ETI) is an emergency procedure performed in civilian and combat casualty care settings to establish an airway. Objective and automated assessment of ETI skills is essential for the training and certification of healthcare providers. However, the current approach is based on manual feedback by an expert, which is subjective, time- and resource-intensive, and is prone to poor inter-rater reliability and halo effects. This work proposes a framework to evaluate ETI skills using single and multi-view videos. The framework consists of two stages. First, a 2D convolutional autoencoder (AE) and a pre-trained self-supervision network extract features from videos. Second, a 1D convolutional enhanced with a cross-view attention module takes the features from the AE as input and outputs predictions for skill evaluation. The ETI datasets were collected in two phases. In the first phase, ETI is performed by two subject cohorts: Experts and Novices. In the second phase, novice subjects perform ETI under time pressure, and the outcome is either Successful or Unsuccessful. A third dataset of videos from a single head-mounted camera for Experts and Novices is also analyzed. The study achieved an accuracy of 100% in identifying Expert/Novice trials in the initial phase. In the second phase, the model showed 85% accuracy in classifying Successful/Unsuccessful procedures. Using head-mounted cameras alone, the model showed a 96% accuracy on Expert and Novice classification while maintaining an accuracy of 85% on classifying successful and unsuccessful. In addition, GradCAMs are presented to explain the differences between Expert and Novice behavior and Successful and Unsuccessful trials. The approach offers a reliable and objective method for automated assessment of ETI skills.

*Keywords: E*ndotracheal intubation, video-based assessment, deep learning, cross-view attention

## 1. Introduction

Endotracheal Intubation (ETI) is an essential airway management procedure that relies on repeated practice and timely intervention for success both in the civilian and combat scenarios [1–3].

---

[1] Corresponding author
Email address: sde@eng.famu.fsu.edu

Notably, combat medics face difficulties in performing ETI, leading to failed airway management to be the second most common cause of death in the battlefield [4,5]. Thus, it is important to develop curriculums to ensure robust evaluation of healthcare providers. The gold standard in ETI skill assessment is Halstedian, i.e., an expert provides real-time or video-based post-hoc feedback to the trainee [6]. This approach has several limitations, including being subjective, manual, time-consuming, and subject to poor inter-rater reliability [2,5].

Recently, video-based assessment (VBA) has received much attention in skill evaluation and education [7–9]. The advantage of VBA is that the experts can prioritize the trainee during sessions while leaving the comprehensive post-hoc feedback later via the video data streams. Furthermore, in recent years, deep neural networks (DNNs) have achieved significant results in video-based tasks, addressing the manual and subjective assessment, especially in related fields such as surgery [10–15]. These frameworks can learn optimal features directly from complex video-data and extract high-level information for classification. However, they are challenging to interpret and fail to provide spatio-temporal feedback. Moreover, these studies utilize a constant camera angle and there is a gap in the literature regarding the utility of multi-view data in skill assessment.

In the literature, multiple techniques have been proposed to analyze single view data. These techniques can learn robust and discriminative features given videos data from a single view. However, their applicability becomes limited when extended to multiple views, as they fail to learn a shared representation of the different viewpoints [16]. Existing works on multiview data can be categorized into two groups [16]. The first group focuses on unsupervised feature extraction from multiple views using variants of auto-encoders [17–19]. They commonly employ unlabeled examples to train a multi-view DNN, then use the network as a feature extractor, followed by a standard classifier. For example, Wang *et al.* [19], analyzed several multi-view techniques involving an autoencoder or a paired feedforward network. They learned representations in which multiple unlabeled views of data are available at training while only one view is available for testing.

The second group of papers proposes to directly build a multi-view DNN for classification [20,21]. For instance, Ainam *et al.* [16] employed a multi-view DNNs that exploits the complementary representation shared between views and proposed a n-pair loss function to better learn a similarity metric. In addition, Chen *et al.* [22] proposed solving the large discrepancy that may exist between extracted features under different views by using an asymmetric distance model. Their network also introduces a cross-view consistency regularization to model the correlation between view-specific features. Similarly, Strijbis *et al.* [23] proposed a multi-view convolutional neural networks (MV-CNNs) for automated eye and tumor segmentation on magnetic resonance imaging (MRI) scan of retinoblastoma patients, and Kan *et al.* [20] proposed a multi-view task agnostic CNN that can be used directly for classification. The approach proposed by Kan et al. involves learning a view-invariant representation using separate view-specific network for each view. However, this method becomes impractical for unbalanced and small datasets, as the limited data may lead to certain branches learning robust features while others suffer from insufficient training.

To overcome these limitations, we propose a framework that can assess clinical skills from entire videos from multiple views of the same procedure. Using cross-view information, instead of relying solely on a single view, is essential in video-based assessment of skills. With multiple camera angles or views, assessors can gain a more comprehensive understanding of the entire procedure. We exploit these differences in viewpoint and propose a pipeline that consists of a 2D autoencoder (AE) and a 1D convolutional classifier. The AE is a convolutional network built on a pre-trained self-supervised model for extracting features from entire videos of different views. The 1D convolutional network with a cross-view attention module takes such features to predict surgical skills. Inspired by the success of attention mechanisms in vision [24,25] and natural language processing[26,27], we propose a cross-view attention (xVA) that exploits the multi-view nature of the data by highlighting the most salient regions in a particular view using masks obtained from a different view. In addition, we provide visual and temporal feedback to the subjects using gradient-based class activation maps (GradCAMs)[28].

The performance of the proposed framework is tested using a dataset comprised of multi-view videos of expert and novice subjects performing ETI procedure on an airway manikin. ETI is a medical procedure for airway management to improve oxygenation in most surgeries. Studies have shown that patients who arrive at a hospital with an ineffective ETI have a lower probability of survival [29,30]. Hence, a proper assessment of ETI skills is fundamental to reducing complications that may increase morbidity and mortality. We tested the efficacy of the model in assessing ETI skill with two classification tasks: (i) a classification analysis to separate the novice subjects from the experts, and (ii) successful and unsuccessful classification of the procedure.

Hence, in this work, we explore deep neural networks for automatic and objective assessment of ETI skills using multi-view video data. We make the following contributions:

1. We propose a framework with a cross-view attention module that can use the full videos from multiple views to provide objective and automated performance evaluation.
2. We provide a visual and temporal heatmap generated via the same DNN pipeline for informative feedback.

This paper is organized as follows: Section 2 provides a review of the related work. Section 3 presents the ETI dataset and the proposed framework. Section 4 details the results. The discussions and conclusions are provided in Sections 5 and 6, respectively.

## 2. Method

We briefly present the dataset used in this paper in Section 3.1 and then describe our method in Section 3.2.

*3.1. Datasets*

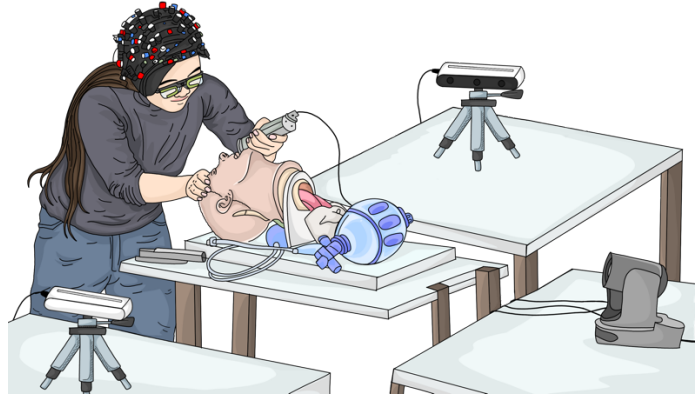

Figure 1: Camera positioning during ETI task to obtain different views of the manikin being intubated.

The videos of the ETI procedure are obtained from an Institutional Review Board (IRB) and Human Research Protection Office (HRPO) approved study at the University at Buffalo where each subject was asked to perform one or more ETI procedures on an airway manikin – Life/form® Airway Larry. There are two distinct datasets that were collected:

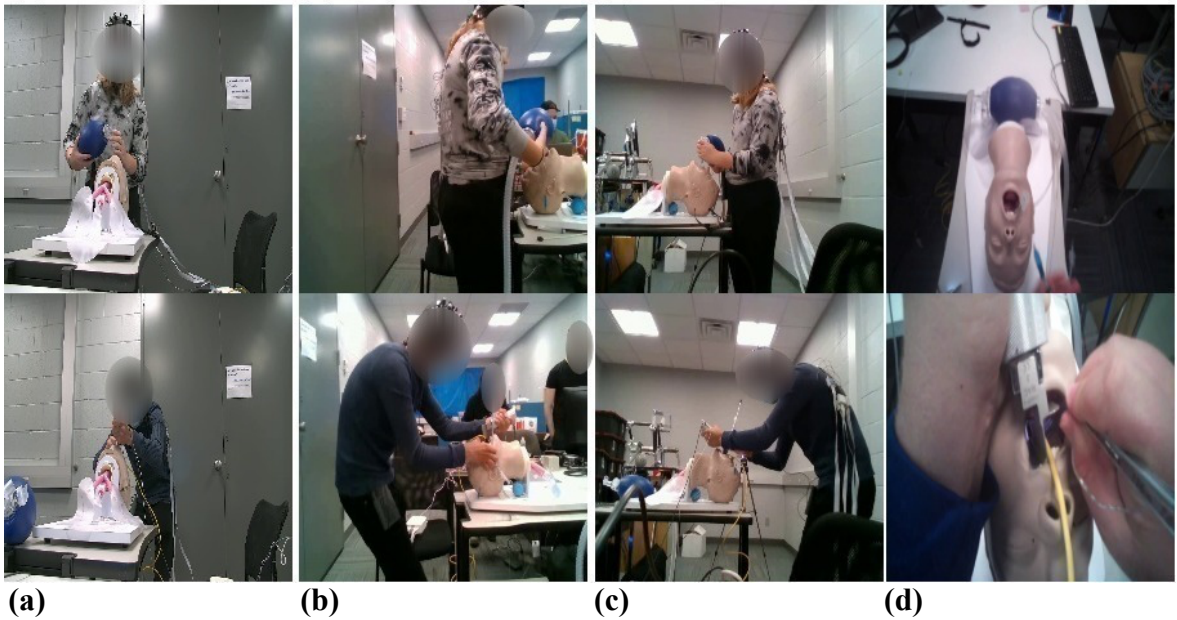

**Figure 2:** Video frames from ETI procedure: **(a)** front, **(b)** left, **(c)** right, and **(d)** head-mounted camera views.

**Time-synchronized multi-view datasets from fixed cameras**: These datasets were obtained using two Intel Realsense side cameras and a PTZOptics front camera at 30 frames per second (Fig 2a-c). The three cameras were time-synchronized and provided fixed views of the scene. Figure 1 shows the positions of the three cameras with respect to the manikin during the procedure The multi-view datasets comprised of two phases of the study:

**Phase 1** dataset consists of the three time-synchronized videos from 17 novice (5 male, 12 female) and 11 expert (7 male, 4 female) subjects. The novice subjects recruited for this study were students in healthcare-related programs with little familiarity with ETI. Experts, on the other hand, had experience ranging from one to over thirty years practicing and teaching the ETI procedure. Each expert and novice subject performed one to five repetitions of the ETI procedure on the airway manikin. Each trial lasted a maximum of three minutes, with two minutes of rest in-between trials. The dataset consists of multi-view videos from 50 successful and 24 unsuccessful trials by novices and 66 successful trials by the expert subjects.

**Phase 2** dataset consists of the three time-synchronized videos from 5 novice subjects (2 females and 3 males) with no overlap with the Phase 1 subjects, with a total of 31 unsuccessful and 106 successful trials. Here, placing the endotracheal tube correctly in the trachea and inflating both lungs within 3 minutes constitute the criteria for being labeled as Successful. While the Phase 1 dataset is used for training/testing the model, the Phase 2 dataset is used to elucidate the model's ability to generalize on an unseen dataset.

**Single-view dataset from a head-mounted camera**: This dataset was obtained using a Tobii Pro Glasses 2 head-mounted camera at 30 frames per second (Fig 2d) from 15 novices (72 trials: 48 successful and 24 unsuccessful trials) and 8 experts (39 successful trials). This dataset is significantly more challenging than the multi-view dataset from the fixed cameras due to head motion and the lack of stability of the videos. However, head-mounted cameras can be easily deployed in simulation centers and for training combat medics, without the need for complex setups involving multiple fixed cameras. Despite their challenges, head-mounted cameras provide essential aspects of the scene in front of the subjects, such as gaze behavior, finding target objects, and visual attention information. Being able to use the head-mounted camera views tests the robustness of our model, though the cross-view attention is not used.

*3.2 Network architecture*

This work aims to predict the outcome of ETI skill assessment using the entire video sequences of the multiple cameras. We seek to harness the temporal and spatial information encoded in the different views to predict ETI skills. To accomplish this task, we consider two networks: a 2D denoising AE and a 1D convolutional classifier.

The encoder $\Phi: \mathbb{R}^n \to \mathbb{R}^{n_z}$ takes as input a set of video frames $Q = \{q^1, q^2, \ldots, q^n\} | q^i \in \mathbb{R}^{c \times h \times w}$ and outputs a set of code vectors $Q_z = \{q_z^1, q_z^2, \ldots, q_z^n\} | q_z^i \in \mathbb{R}^{n_z}$. Here, $c \times h \times w$ is the spatial dimension of the input frame and $n_z$ is the length of the code vector. The decoder $\Psi: \mathbb{R}^{n_z} \to \mathbb{R}^n$ then takes the input frame $q^i$ that is coded as $q_z^i$ and outputs the reconstructed frame $\tilde{q}^i$. We also use a pre-trained self-supervision model [31] that exposes the inner structure of the input data and constrains the decoder to output data that resembles the input data with low reconstruction loss. The pre-trained self-supervision model introduces pre-specified pixels that must be present in the output data. This also incorporates auxiliary knowledge into the model without requiring any modification of the network parameters.

The AE is trained to minimize the reconstruction loss between the input and the output and is defined as follows:

$$\mathcal{L}_{DAE} = \frac{1}{N}\sum_{i=1}^{N}\left\|q^i - \tilde{q}^i\right\|_2^2$$

where $N$ is the size of the mini-batch.

The AE is enhanced with the perceptual loss [32] using the features extracted from the self-supervision model and is expressed as follows:

$$\mathcal{L}_{perc}^{\phi,i}(q^i, \tilde{q}^i) = \frac{1}{C_i H_i W_i}\left\|\phi(q^i) - \phi(\tilde{q}^i)\right\|_2^2,$$

where $C_i, H_i, W_i$ are the channel, height, and width, respectively, and $\phi$ is the self-supervised model. The input to $\phi$ is a reconstructed image from the decoder and the original high-quality image. We take advantage of this loss to capture more semantic information while guiding our model to generalize. The final loss is then defined as:

$$\mathcal{L}_{final\_AE} = \mathcal{L}_{DAE} + \mathcal{L}_{perc}.$$

**Table 1** shows the detailed architecture of the network, and **Figure 3** shows the overall framework.

**Table 1:** The 2D convolutional AE network structure. After each Conv2d layer, we used a SELU activation function.

| Name | Input size | Kernel size | (Padding, Stride) | Output size |
|---|---|---|---|---|
| encoder.conv2d_1 | [3 × 256 × 256] | (3 × 3) | 1, 2 | [32 × 128× 128] |
| encoder.conv2d_2 | [32 × 128× 128] | (3 × 3) | 1, 1 | [32 × 128× 128] |
| encoder.conv2d_3 | [32 × 128× 128] | (3 × 3) | 1, 2 | [64 × 64 × 64] |
| encoder.conv2d_4 | [64 × 64 × 64] | (3 × 3) | 1, 2 | [64 × 32 × 32] |
| encoder.conv2d_5 | [64 × 32 × 32] | (3 × 3) | 1, 2 | [128 × 16 × 16] |
| encoder.conv2d_6 | [128 × 16 × 16] | (3 × 3) | 1, 2 | [128 × 8 × 8] |
| encoder.conv2d_7 | [128 × 8 × 8] | (3 × 3) | 1, 2 | [128 × 4 × 4] |
| encoder.conv2d_8 | [128 × 4 × 4] | (3 × 3) | 1, 2 | [128 × 2 × 2] |
| encoder.GAP2d | [128 × 2 × 2] | (1 × 1) | - | [128 × 1 × 1] |
| decoder.linear | [128] | - | - | [512] |
| decoder.convTrans2d_1 | [128 × 2 × 2] | (3 × 3) | 1, 2 | [128 × 4 × 4] |
| decoder.convTrans2d_2 | [128 × 4 × 4] | (3 × 3) | 1, 2 | [128 × 8 × 8] |
| decoder.convTrans2d_3 | [128 × 8 × 8] | (3 × 3) | 1, 2 | [64 × 16 × 16] |
| decoder.convTrans2d_4 | [64 × 16 × 16] | (3 × 3) | 1, 2 | [64 × 32 × 32] |
| decoder.convTrans2d_5 | [64 × 32 × 32] | (3 × 3) | 1, 2 | [64 × 64 × 64] |
| decoder.convTrans2d_6 | [64 × 64 × 64] | (3 × 3) | 1, 2 | [32 × 128 × 128] |
| decoder.convTrans2d_7 | [32 × 128 × 128] | (3 × 3) | 1, 2 | [3 × 256 × 256] |

The architecture of the encoder consists of eight convolutional blocks (encoder.conv2d_1 to encoder.conv2d_8). Each 2D convolution operator slides a kernel of weight over the image data and performs element-wise multiplication with the data that falls under the kernel and can be precisely described as:

$$out\left(N_i, C_{out_j}\right) = bias\left(C_{out_j}\right) + \sum_{k=0}^{C_{in}-1} weight\left(C_{out_j}, k\right) * input(N_i, k),$$

where $N$ is the batch size, $C$ denotes the number of channels ($C = 3$ for RGB images), $H$ is the height of input, and $W$ is the width.

Similarly, the architecture of the decoder contains seven deconvolutional blocks (decoder.convTrans2d_1 to decoder.convTrans2d_7). Each block applies a 2D deconvolution operator (*a.k.a* transposed convolution operator) over the input. When initialized with the same parameters, encoder.conv2d_8 and decoder.convTrans2d_1 are inverses of each other in regard to the input and output shapes. Between the encoder and the decoder, we use a 2D global average pooling (encoder.GAP2d) to down-sample the input followed by a linear transformation (decoder.linear).

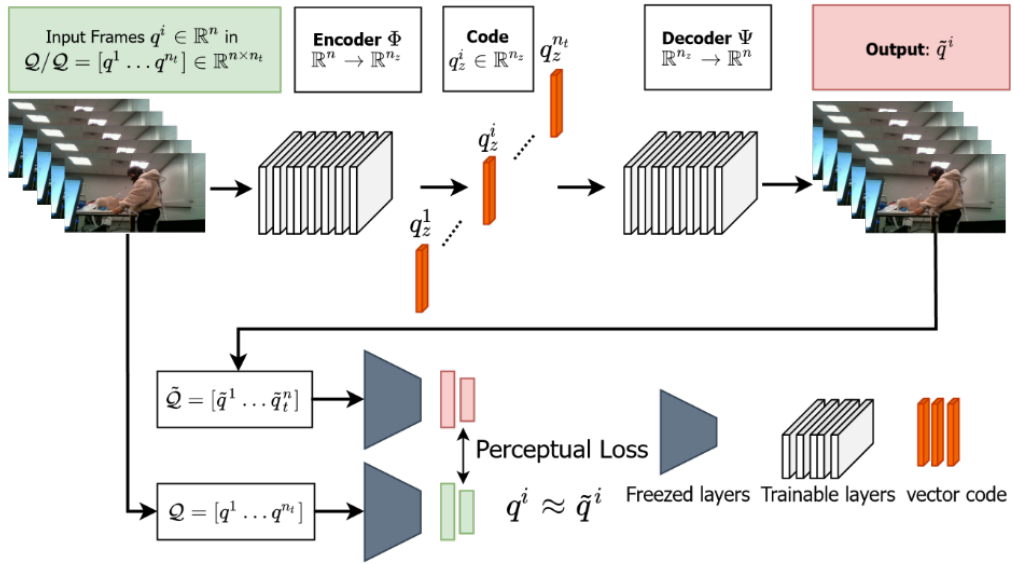

**Figure 3:** The AE framework with the self-supervised model.

After extracting video features from the AE, we use a 1D convolutional network for classification (**Figure 4**). It includes Conv1D layers, the proposed cross-view attention (xVA) layer, and squeeze-and-excitation layers (SE layers). The SE layers work as channel-wise attention based on Hu *et al.* [33]'s implementation.

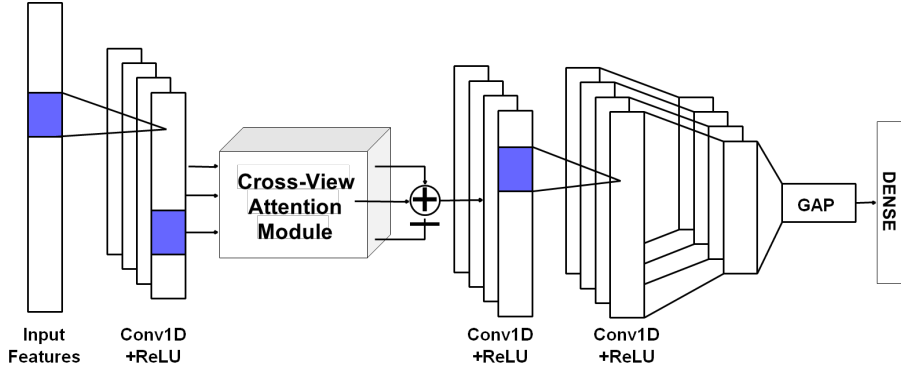

**Figure 4:** Conv1D classifier architecture. SE (Squeeze-and-Excitation network [33]) is included after each block (Conv1D + ReLU).

The proposed cross-view attention is based on feature fusion. It is a fusion method where attention masks from one view are used to highlight the extracted features in another view. The cross-view attention is different from the self-attention mechanism. In self-attention, masks are used to highlight their features. Due to the nature of the dataset, a cross-view attention that leverages multiple views is necessary. Our cross-view attention operates as follows. Given two input features $(v_1, v_2) \in R^{T \times C_{in}}$ with channels $C_{in}$, and temporal length $T$, the output $y \in R^{T \times C_{out}}$ is computed as:

$$y = \text{concat}([o_1, o_2]) \in R^{T \times C_{out}},$$

*such that*

$$o_1 = [m_1 \odot v_2] * v_1,$$

$$o_2 = [m_2 \odot v_1] * v_2,$$

$$m_1 = \sigma (v_1 \odot v_2),$$

$$m_2 = \sigma (v_2 \odot v_1),$$

where σ is the SoftMax operator, $\odot$ is the dot product, $*$ is the matrix multiplication.

The 1D convolutional network is trained using a cross-entropy loss:

$$L_{xent} = -\frac{1}{m}\sum_{i=1}^{m} y_i \log(\hat{y}_i),$$

where $y_i$ is the ground-truth label, $m$ is the number of classes, and $\hat{y}_i$ is the predicted label.

**Baseline network:** In order to evaluate single view data, we present a baseline network that incorporates all components of our proposed network, excluding the cross-view attention mechanism. In other words, the baseline network takes video input from a single view and predicts the outcome. This baseline network individually assesses the left and right views and the head-mounted video.

*3.3. Implementation details*

The framework was implemented using PyTorch library [34]. For the AE, we used a pre-trained SimCLR [31] as the self-supervision model and fine-tuned it on our dataset. The image frames were extracted from the videos and resized to $256 \times 256$. The pixels were scaled between -1 and 1 and gaussian noise was added to the AE input. The AE was trained for 100 epochs using Adam optimizer [35] on a batch size of 128 and with the default hyperparameters [35]. The SimCLR model parameters were frozen during the training process. To further improve the training capability, the learning rate was gradually decreased by a factor of 0.2 every 20 epochs or once learning stagnated, to a minimum value of $5e - 5$.

The 1D convolutional classifier was trained on a dataset comprised of variable length sequences of a feature vector of size 32 ($i.e., n_z = 32$) representing each frame in the video. The classifier was trained for 50 epochs using the Adam optimizer on a unit batch size with default hyperparameters [35]. An independent assessment of the classifier was performed using a 10-fold cross-validation approach.

To compute the temporal GradCAM, we taped into the last convolution layer of the Conv1D (before the SoftMax) network to compute the gradient of the score for the given class with respect to the features maps. Each temporal location *i* in the class-specific saliency map $L^c$ is calculated as:

$$L_i^c = \sum_k w_k^c \cdot A_i^k,$$

where the weight $w_k^c = \frac{1}{n} \sum_i \frac{\delta Y^c}{\delta A_i^k}$, $n$ is the sequence length and $A^k$ is the feature map. $L_i^c$ directly correlate with the importance of a particular temporal location (*i*) for a particular class c and thus functions as a temporal explanation of the class predicted by the network.

3. **Results**

In section 3.1, we provide results using the baseline network, i.e., using each view of the time-synchronized multi-view datasets independently without cross-view attention. In sections 3.2, we introduce cross-view attention for the same datasets. Finally, in section 3.3, we use our model, without the cross-view attention for the single-view dataset using the head-mounted camera.

*3.1. Classification tasks using single camera views of the time-synchronized multi-view datasets for Phase 1.*

Table 2 shows the classification results of the three views trained and tested on the Phase 1 dataset. We obtained an accuracy of 0.84, 0.85, and 0.72 on the successful/unsuccessful task using left, right and front views, respectively. On the expert/novice task, we achieve much higher accuracies of 0.97, 0.94, and 0.98, respectively. These results on single views demonstrate the network's ability to learn discriminative features. However, the obtained performances remain modest compared to multi-view, as indicated in Tables 3, 4, and 5. Although single views achieve interesting results, they will not provide assessors with a comprehensive evaluation. We believe that cross-view information will allow not only for improved error detection, but it will also allow assessors to identify and analyze deviations in technique by comparing the different camera angles. Consequently, to maintain accuracy classification when using multiple views, we have introduced cross-view attention as a mechanism to capture and leverage the relationships between the three views.

Table 2: Classification result using views separately. Suc.vs.Uns and Exp.vs.Nov stands for successful/Unsuccessful and expert/novice tasks, respectively.

| Metrics | Left view | | Right view | | Front view | |
|---|---|---|---|---|---|---|
| | Suc.vs.Uns | Exp.vs.Nov | Suc.vs.Uns | Exp.vs.Nov | Suc.vs.Uns | Exp.vs.Nov |
| Accuracy | 0.84 | 0.97 | 0.85 | 0.94 | 0.72 | 0.98 |
| Sensitivity | 0.65 | 1.0 | 0.65 | 1.0 | 0.28 | 0.97 |
| Specificity | 0.93 | 0.93 | 0.95 | 0.82 | 0.94 | 1.0 |
| F1 score | 0.73 | 0.98 | 0.75 | 0.95 | 0.40 | 0.98 |
| ROC AUC | 0.85 | 0.99 | 0.88 | 0.98 | 0.67 | 0.99 |

*3.2. Classification tasks using all three fixed cameras for Phase 1 and 2 datasets*

First, we present results for the Phase 1 dataset. **Table 3** reports the classification results for the successful vs. unsuccessful classification.

**Table 3:** Results for Successful/Unsuccessful classification using the three time-synchronized fixed cameras for Phase 1 data.

| Successful/Unsuccessful classification | | w/o xVA | w/o SE |
|---|---|---|---|
| **Accuracy** | **0.83** | 0.74 | 0.77 |
| Sensitivity | 0.77 | 0.63 | 0.50 |
| Specificity | 0.86 | 0.79 | 0.91 |
| F1 score | 0.76 | 0.62 | 0.59 |
| ROC AUC | 0.86 | 0.84 | 0.87 |
| MCC | 0.63 | 0.43 | 0.46 |

The findings reveal that the model detects the trials with an accuracy of 0.83. The model is balanced (F1 = 0.76) in that it detects successful samples with a sensitivity performance of 0.77 while also avoiding unsuccessful ETI trials from being labeled as successful (specificity = 0.86). The receiver operator characteristics (ROC) curve and the precision-recall (PR) curve have area under the curve (AUC) values of 0.86 and 0.89, respectively, demonstrating the efficiency of our approach in distinguishing successful from unsuccessful trials.

Furthermore, we analyze the contribution of the attention modules, i.e., xVA and SE, to the skill classification. As seen in **Table 3**, the attention modules improved the accuracy via recalibrating the salient features (accuracy = 0.83 / 7.2% higher and F1 = 0.76 / 22.4% higher) and prevented overfitting to the unsuccessful samples as seen in sensitivity without the xVA and SE (18.2% and 35.1% drop), even though the separability is marginally lower (ROC AUC = 0.82 / 3.5% lower). This is possibly due to the overlapping distributions, which makes it harder for the classifier to differentiate the two classes, or the class imbalance, which causes the model to capture less discriminatory information for the smaller distribution. However, overall, we detect a performance improvement.

**Table 4** lists performance metrics for classifying experts and novices. We see that the model identifies all the expert and novice trials of ETI task accurately. Furthermore, the AUC value of 1.0 for the ROC and PR curves demonstrates the model's ability to separate classes with a greater margin. There is no performance loss without the attention modules.

**Table 4:** Results for Expert/Novice classification using the three time-synchronized fixed cameras for Phase 1 data.

| Expert/Novice classification | | w/o xVA | w/o SE |
|---|---|---|---|
| **Accuracy** | **1.0** | 1.0 | 1.0 |
| Sensitivity | 1.0 | 1.0 | 1.0 |
| Specificity | 1.0 | 1.0 | 1.0 |
| F1 score | 1.0 | 1.0 | 1.0 |
| ROC AUC | 1.0 | 1.0 | 1.0 |

To further investigate the model's ability to generalize to new and previously unseen dataset, we use data collected during Phase 2 study. First, we used the AE as a feature extractor. AE takes raw video from Phase 2 as input and output feature vectors. Then, the Conv1D classifier network is fined-tuned on the extracted features to predict the trial's outcome. Here, we no longer trained AE on Phase 2 data, but use the AE model previously trained on Phase 1 data. The results of this evaluation are shown in **Table 5**.

**Table 5:** Results for Successful/Unsuccessful classification using the three time-synchronized fixed cameras for Phase 2 data.

| Successful/Unsuccessful classification | |
|---|---|
| Accuracy | 0.92 |
| Sensitivity | 0.71 |
| Specificity | 0.98 |
| F1 score | 0.80 |
| ROC AUC | 0.93 |
| MCC | 0.76 |

Using 31 unsuccessful and 106 successful trials from Phase 2 study, we achieved a high accuracy score of 0.92 and an F1 score of 0.8, indicating perfect precision and recall. Notice that classification accuracy is better for Phase 2 data. We hypothesize that the enhanced accuracy is associated with a larger Phase 2 dataset. To substantiate this claim, we randomly sampled 50 successful and 22 unsuccessful trials from the 106 successful and 31 unsuccessful trials from Phase 2 to match those of Phase 1. We report the average results after ten runs in **Table 6**.

**Table 6:** Successful/Unsuccessful classification results using a subset of Phase 2 data. (STD = 0.66)

| Successful/Unsuccessful classification | | |
|---|---|---|
| **Metrics** | **Phase 2** | **Phase 1** |
| Accuracy | 0.85 | 0.83 |
| Sensitivity | 0.62 | 0.77 |
| Specificity | 0.98 | 0.86 |
| F1 score | 0.75 | 0.76 |
| ROC AUC | 0.89 | 0.82 |
| MCC | 0.67 | 0.63 |

With comparable sample sizes, the model predictions are consistent between the two datasets, as illustrated in **Table 6**. We achieved an accuracy of 0.85 on Phase 2 data, which is comparable of the accuracy obtained in Phase 1. Moreover, all the results of the ten runs are close to the reported average as indicated by the low standard deviation value of 0.066.

### 3.3. Classification tasks using the head-mounted camera

The performance results for expert/novice and successful/unsuccessful classification tasks using videos from the head mounted camera are presented in **Table 7**. The high accuracy of 0.96 and the Matthews correlation coefficient (MCC) of 0.92 highlights the ability of our model to differentiate between expert and novice trials, despite the imbalance in the dataset. The network is shown to classify the unsuccessful and successful trials with accuracy of 0.78, which is comparable to that of the classification using single static camera views, see Table 2

**Table 7:** Classification results using the head-mounted camera view.

| Expert/Novice classification | | Successful/Unsuccessful classification | |
|---|---|---|---|
| **Accuracy** | **0.96** | **Accuracy** | **0.85** |
| Sensitivity | 0.92 | Sensitivity | 0.72 |
| Specificity | 0.99 | Specificity | 0.88 |
| F1 score | 0.95 | F1 score | 0.68 |
| MCC | 0.92 | MCC | 0.58 |

**Figure 5** shows the ROC and PR curves with the AUC measuring the degree of separability or network confidence in separating the two classes. To measure the degree of separability or network confidence in separating the two classes, we report the ROC AUC and PR AUC of the two classifications. The Expert/Novice task performs well, with an AUC of 0.99 for ROC and 0.98 for PR. Similarly, with an AUC of 0.82 for ROC and 0.67 for PR. These show the ability of our classifier to distinguish between classes. In addition, we show the trade-off between sensitivity (TPR) and specificity (1 - FPR). The results still indicate a better performance for the two tasks using head-mounted videos alone.

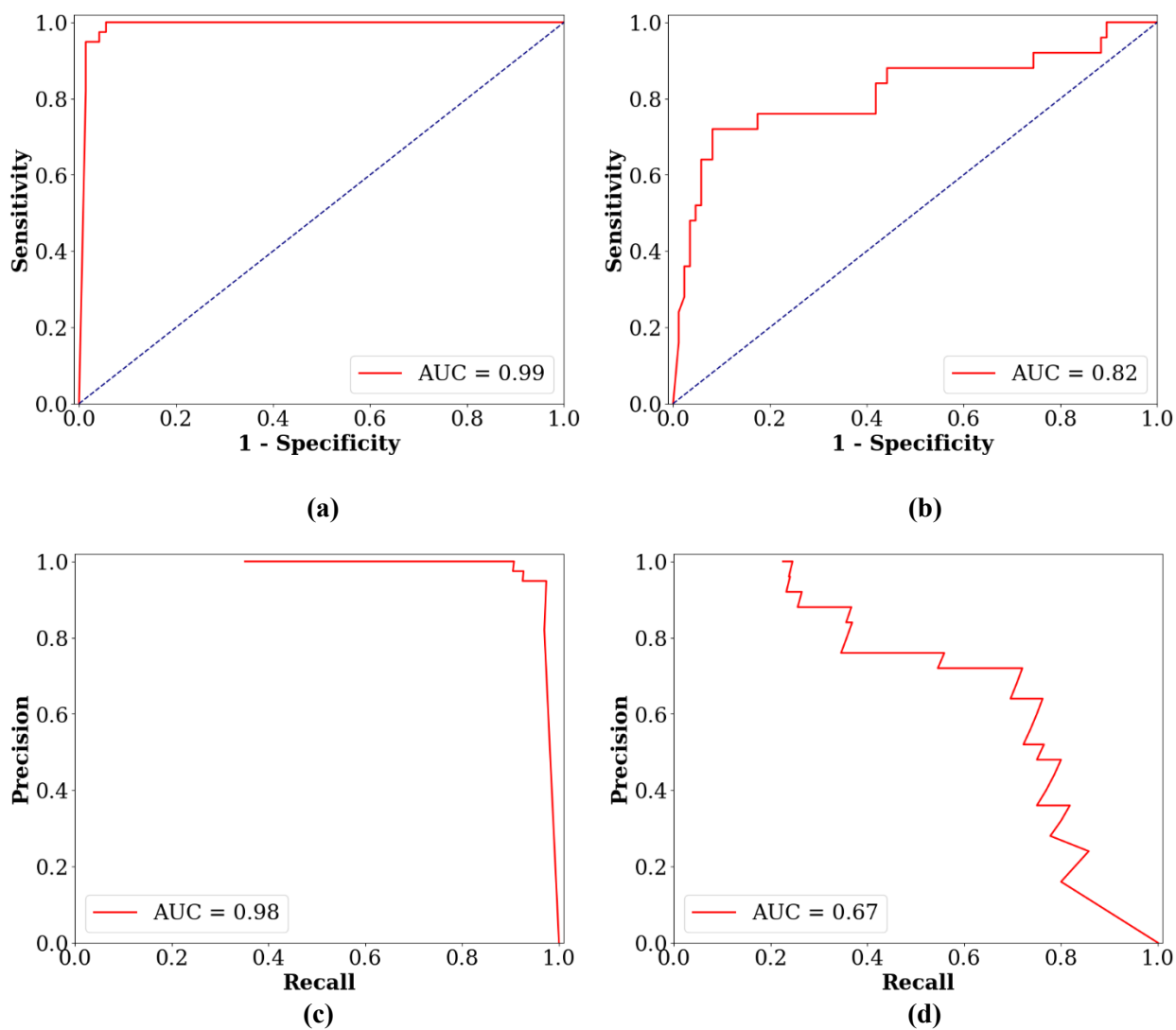

**Figure 5:** ROC (top) and PR Curves (bottom) and corresponding AUC values. (a), (c) are for expert/novice and (b)-(d) for unsuccessful/successful.

## 4. Discussion

In this work, we have developed a deep learning model to assess ETI task performance. Such a model provides objective measures of the performance of clinicians and field medics, eliminating potential biases and inconsistencies in the evaluation process. However, to be acceptable to the clinical community, the model must be trustworthy, i.e., there must be confidence in its predictions. Also, a good model must be able to provide clinically relevant feedback to the trainee on their performance, without the need for manual supervision and debriefing. Trustworthiness and feedback are discussed in Sections 4.1 and 4.2, respectively.

## 4.1. Trustworthiness of the model

We gauge the trustworthiness of our model based on metrics proposed in [36,37]. These metrics are developed based on the SoftMax probability of the predictions. The model is considered more reliable for a binary classification study when the SoftMax probabilities, i.e., collectively trust spectrum, are farther away from the threshold (0.5). **Figure 6** shows the trust spectrum, i.e., the density of SoftMax probability per sample and the corresponding area under curve values, namely NetTrustScore (NTS).

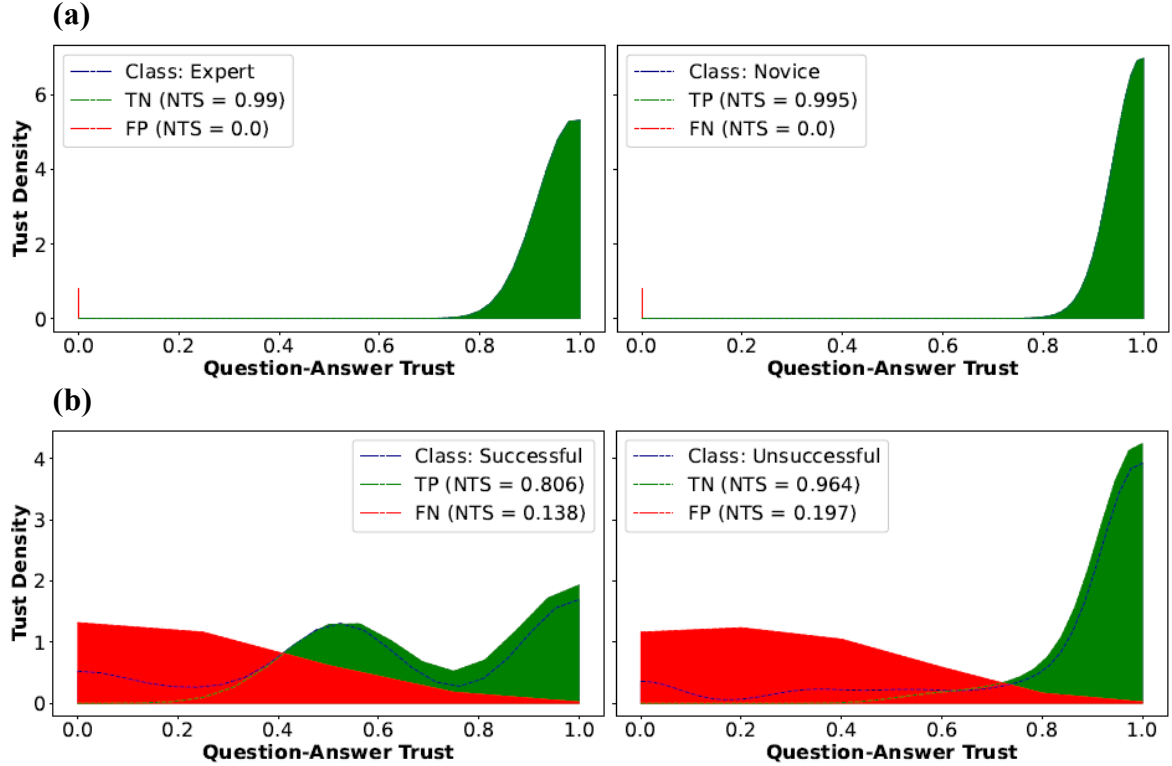

**Figure 6:** Trustworthiness spectrum and NTS for **(a)** Expert vs. Novice and **(b)** Successful vs. Unsuccessful classification.

The figure shows that our model has high NTS scores (> 0.8) for true predictions, i.e., the spectrum is skewed towards the right, where higher SoftMax probabilities are represented for both successful and unsuccessful cases. This indicates that the model has a robust decision criterion for the true predictions, enhancing its reliability. However, the model has a trust spectrum farther away from the threshold of 0.5 for false prediction in both classes, resulting NTSs less than 0.2. This signifies that the model can benefit from additional data on the ETI task [36].

The trustworthiness of our model for the expert vs. novice classification of ETI task via moving head mounted camera view, yields NTS scores of 0.985 and 0.979, respectively, for the TP and TN cases (**Figure 7**), which is comparable to that for the stable camera views (TP = 0.995; TN = 0.990; Fig. 8). Notably, for the failed predictions, the NTS scores are very close to the threshold (0.5) signifying that the model does not have a strong opinion on the false predictions, a desired trait towards improving the results. However, it is important to mention that there is only one FP and FN sample. Therefore, to solidify the reliability of the model, i.e., to show that the spectrum is closer to the threshold for false predictions, more data are needed.

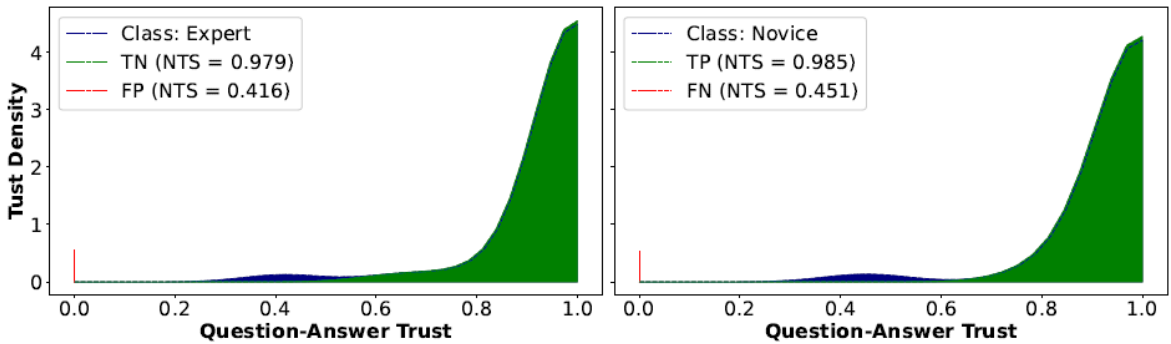

**Figure 7:** Trustworthiness spectrum and NTS for Expert vs. Novice using head-mounted video.

*4.2. Feedback*

Feedback is provided using both spatial and temporal heatmaps generated using GradCAM[28]. GradCAM helps visualize parts of the ETI task that contributes the most to the classification. It generates heatmaps that highlight the regions of the input that are important for a machine learning model's prediction. In the context of ETI procedure, GradCAM is used to generate heatmaps that highlight the regions of a video input stream that are important for the successful placement of an endotracheal tube. By analyzing the heatmaps generated by GradCAM, clinicians can gain insights into the factors that contribute to successful intubation. For example, the heatmaps may show that certain features, such head position, hand movements, even vocal cords, tracheal rings, are particularly important for accurate tube placement. This information can help clinicians to identify good posture or areas of focus during the intubation procedure and may improve the overall success rate of intubation.

Additionally, these heatmaps can be used to train and refine machine learning models that are designed to assist with endotracheal intubation. By using the heatmaps as a form of feedback, researchers can identify the features that are most relevant, and can develop algorithms that are optimized for these features. This may lead to more accurate and reliable automated intubation systems in the future. Heatmaps generated using GradCAM are shown in Figure 8-11. The heatmaps highlights the segments of the video that the network is focusing on to make a prediction.

As shown in **Figure 8**, the pose of an expert and a novice could help identify correct (or incorrect) movements that result in a successful (or unsuccessful) procedure. The expert in Figure 8b maintains a lowered/squatting position during the entire intubation procedure (frames 2 and 3), whereas the novice in Figure 8a crouches at the start (frame 2), but then stands toward the end. Proper positioning is required to obtain a good view of the airway for successful intubation. Standing prior to the end of the task may cause the provider to lose sight of the airway. The spatial heatmaps can also pick up differences in hand pose. Choking down on the laryngoscope blade (near the handle-blade connection) can lead to using the wrist as leverage, torquing the blade onto the teeth and chipping the tooth. While the provider may successfully intubate, the patient may be harmed. The heatmap for the expert shows that the hand is placed more at the center of the handle and the elbow is rested on the table to allow the provider to use the elbow and shoulder as the fulcrum to lift the mandible and view the vocal cords.

In **Figure 9**, we study ETI task sequences for a novice and an expert subject and visualize the temporal GradCAM and the frames associated with the activations. In this heatmap, the x-axis corresponds to the discrete sequence frames of the input video, while the y-axis represents the intensity or magnitude of the assigned weights. Hence, high values on the y-axis indicate that the frames in the input data have received higher weights, indicating their significance in the overall processing. **Figure 9** shows that the model could detect the time sequences that contribute to the classification of the task. This feedback could explain which movements differentiate novice subjects from experts. Overall, for the novices, the weights are elevated throughout the entire procedure. The novices spend much of the task duration attempting to identify the airway and place the endotracheal tube. When the tube is perceived as in the trachea, the weights start to decrease. With the expert, the weights are primarily elevated at the beginning of the task during the period of identifying landmarks. Once the vocal cords are identified, the temporal heatmap weights decrease. Identifying the appropriate landmarks and vocal cords is one of the most important and challenging parts of the task. Using spatial heatmaps, we can provide constructive feedback to trainees and help them improve their performance.

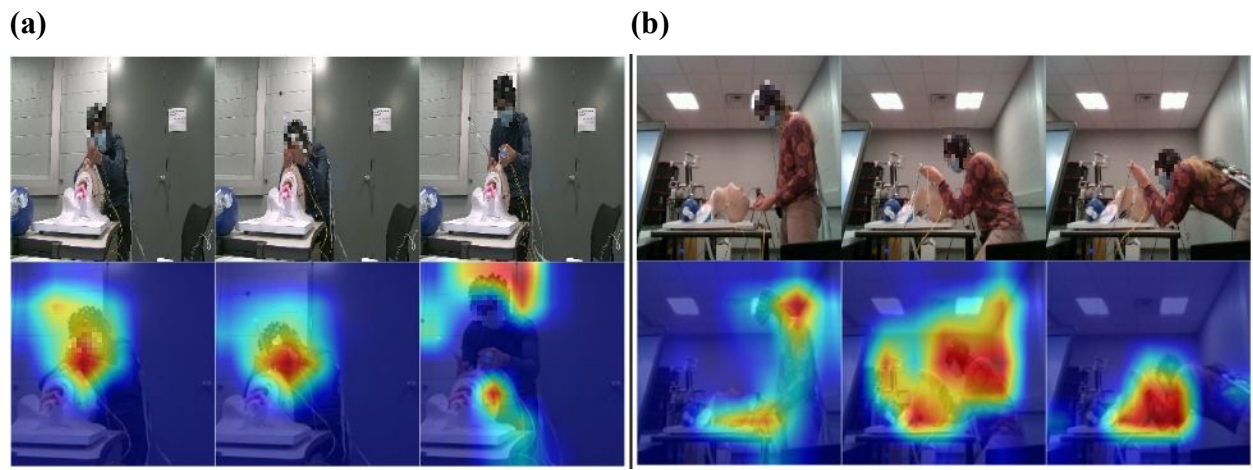

**Figure 8:** Spatial heatmap provided by GradCAM. **(a)** shows a Novice and their corresponding heatmap. **(b)** represents an Expert and the corresponding heatmaps.

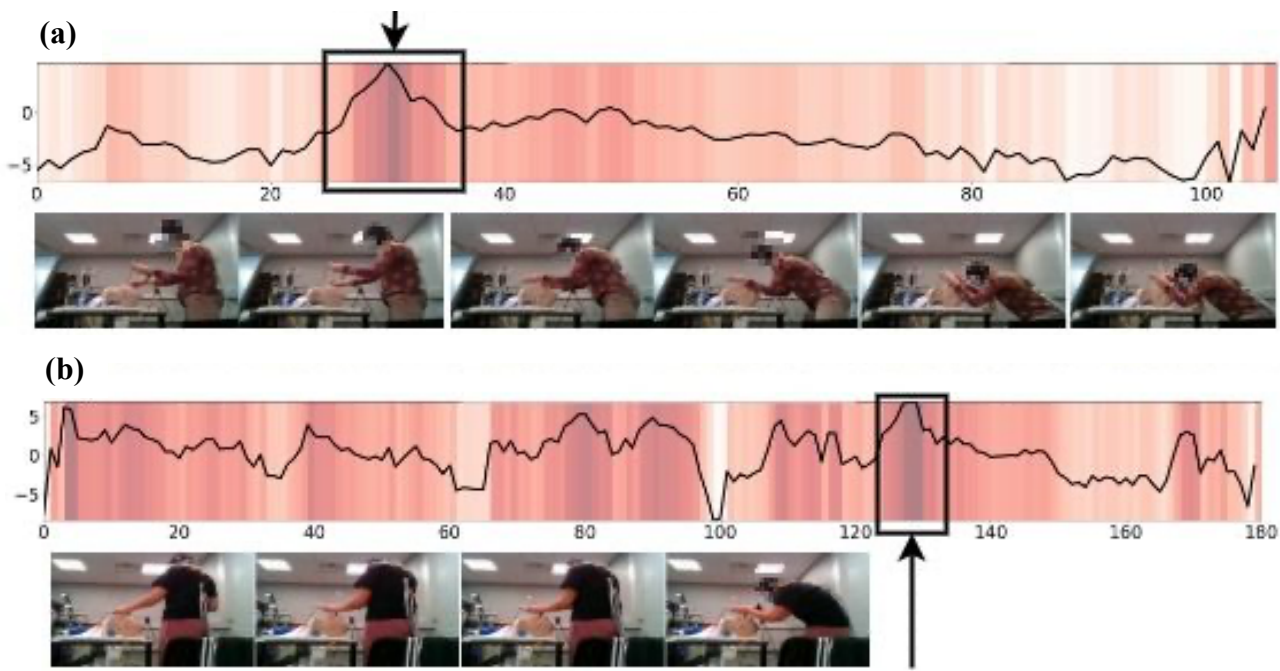

**Figure 9:** Temporal heatmaps for **(a)** Expert and **(b)** Novice subjects and their corresponding frames.

We also studied the temporal heatmaps for the head-mounted cameras. For novices (see **Figure 10**), the network highlights parts of the video where the subject stands on the patient's right side and does not see the patient's head. As a result, standing sideways while performing the ETI task is an indicator used by the network to differentiate novices from experts. Moreover, unnecessary objects such as the end of the subject's sleeve that occludes the airway manikin's face is another indicator captured by the network. On the other hand, experts stand near the top, facing the manikin's head, and can see the airways. There is also an elevated weight from the model when the expert is checking the equipment, which is a task component stressed during training. When comparing repeat attempts for the novices, after successful intubations, the participants adjusted their body position and the temporal heatmap weights shifted such that the elevated period is near the start and declines quickly after successful tube placement.

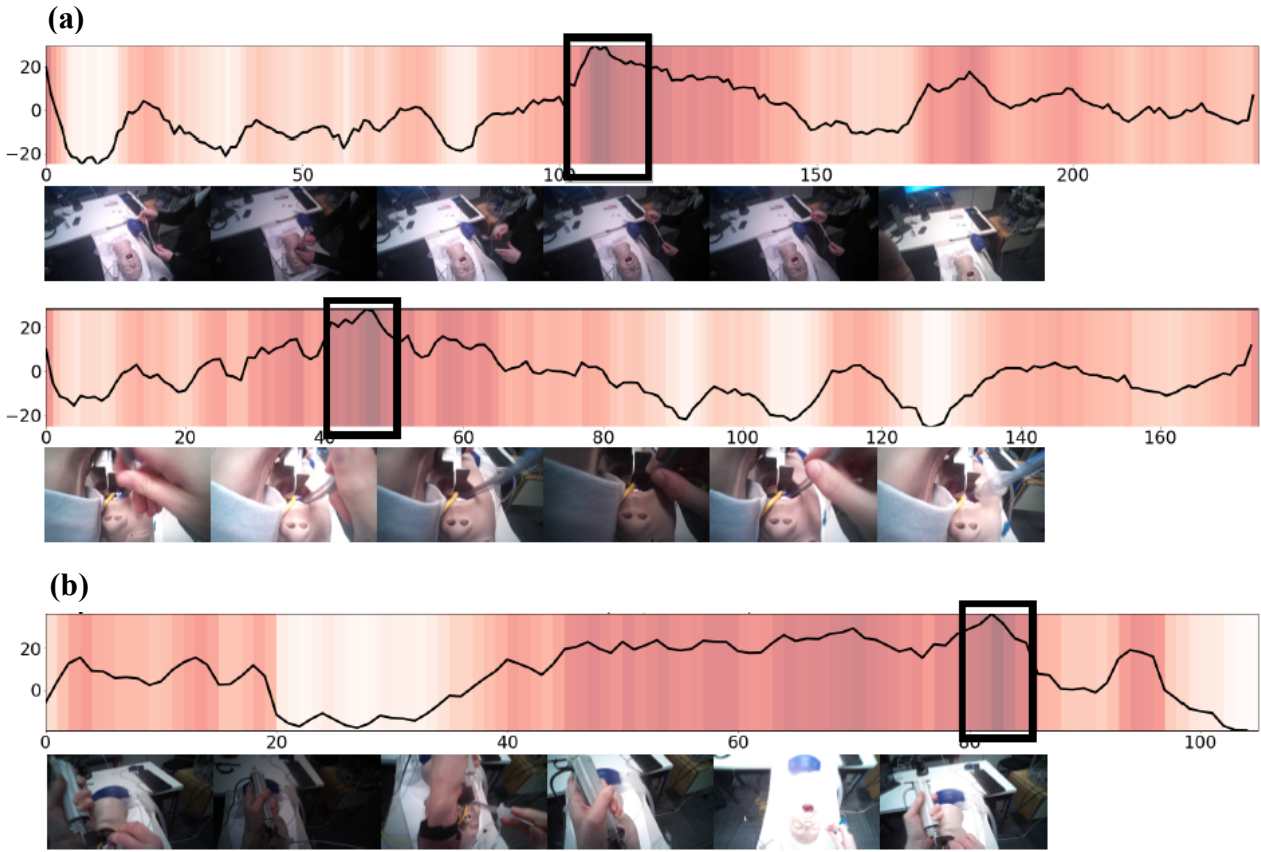

**Figure 10:** Temporal heatmaps for Expert **(a)** and Novice **(b)** subjects using head-mounted video.

Similarly, **Figure 11** shows the indicators used by the network to differentiate successful from unsuccessful trials. For successful trials, the forehead is well exposed, and the airway is visible, whereas for unsuccessful trials, the subject is seen to apply pressure on the manikin forehead.

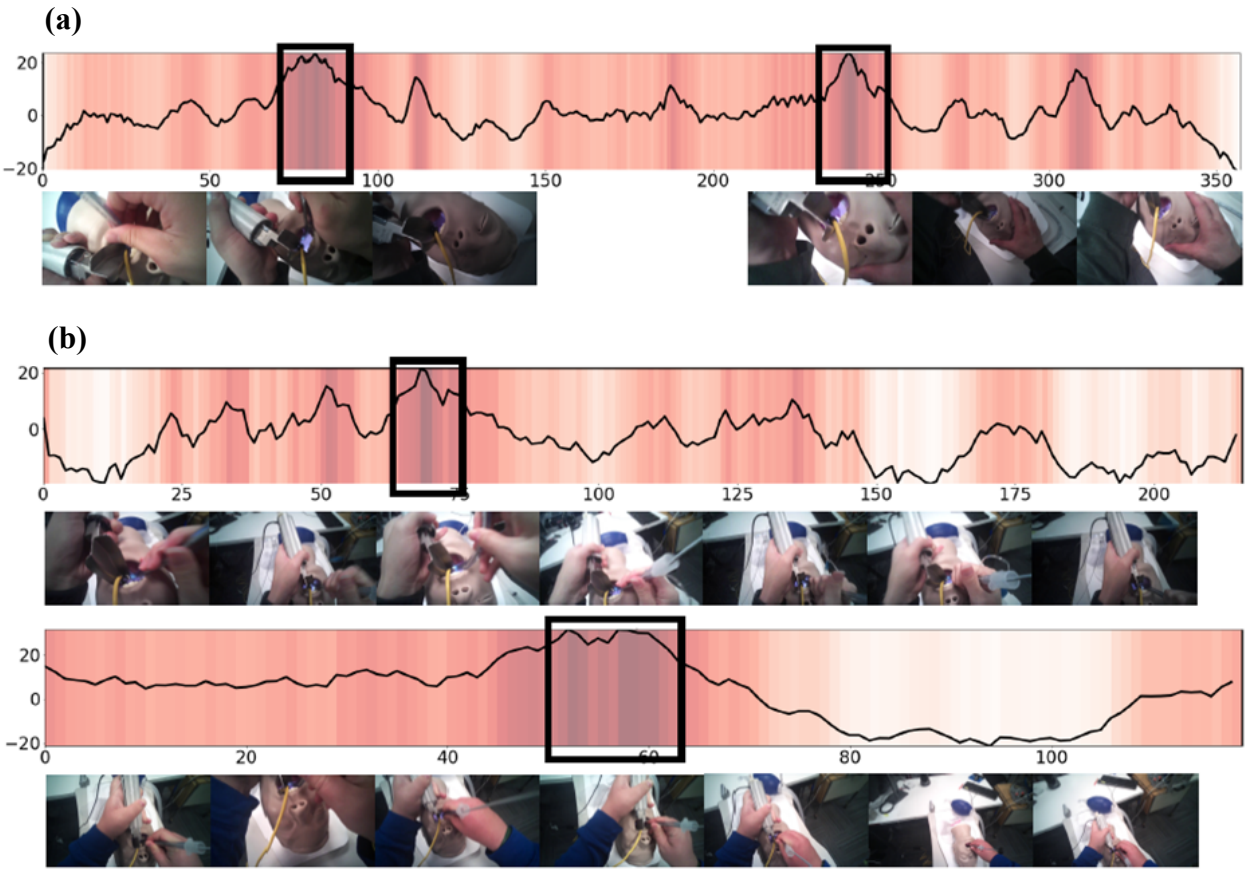

**Figure 11:** Temporal heatmap for **(a)** Unsuccessful and **(b)** Successful trial using head-mounted videos. What are being shown in the box?

## 5. Conclusion

This paper presents a deep learning model to assess ETI skills using videos. Results are presented for multi-camera views that are synchronized, as well as single camera views including a head-mounted camera. The model is shown to generalize to unseen datasets. It is also reasonable trustworthy and provides valuable interpretable feedback based on spatio-temporal heatmaps.

However, the modeling pipeline is not fully automated and consists of two steps: a 2D autoencoder for feature extraction and a Conv1D network for classification. In future work, we aim to develop an end-to-end deep learning model to assess skills while continuing to provide interpretable feedback. Moreover, the manikin lacks realism which may have affected the participant's performance negatively. Similarly, the standard size of the manikin does not represent the diversity in real-life, limiting our results to a specific distribution. Being able to demonstrate the effectiveness of the model for intubation tasks in actual patients will be another goal of this study.

**Acknowledgment**

We gratefully acknowledge the support of this work through the U.S. Army Futures Command, Combat Capabilities Development Command Soldier Center STTC cooperative research agreement #W912CG-21-2-0001. The authors would like to thank Dr. Lora Cavuoto and her group for data collection and the attending subjects for their dedication to this study.

**Data and Code availability**

All relevant data (two-dimensional feature: Temporal and feature vector, 1D feature vectors, and computer code) are available via the authors' GitHub repository at https://github.com/jpainam/eti

**References**

[1] C. Lim, H.S. Ko, S. Cho, I. Ohu, H.E. Wang, R. Griffin, B. Kerrey, J.N. Carlson, Development of a Hand Motion-based Assessment System for Endotracheal Intubation Training., J Med Syst. 45 (2021) 81. https://doi.org/10.1007/s10916-021-01755-2.

[2] S. Zhao, W. Li, X. Zhang, X. Xiao, Y. Meng, J. Philbeck, N. Younes, R. Alahmadi, L. Soghier, J. Hahn, Automated Assessment System with Cross Reality for Neonatal Endotracheal Intubation Training, in: 2020 IEEE Conference on Virtual Reality and 3D User Interfaces Abstracts and Workshops (VRW), 2020: pp. 738–739. https://doi.org/10.1109/VRW50115.2020.00220.

[3] C.N. Bedolla, Airway Management: Review of Current Devices and Development of a Novel Endotracheal Tube For Emergency Combat Care, The University of Texas at San Antonio, 2022.

[4] S. Zhao, X. Xiao, X. Zhang, W.L. Yan Meng, L. Soghier, J.K. Hahn, Automated Assessment System for Neonatal Endotracheal Intubation Using Dilated Convolutional Neural Network., Annu Int Conf IEEE Eng Med Biol Soc. 2020 (2020) 5455–5458. https://doi.org/10.1109/EMBC44109.2020.9176329.

[5] X. Xiao, S. Zhao, X. Zhang, L. Soghier, J. Hahn, Automated Assessment of Neonatal Endotracheal Intubation Measured by a Virtual Reality Simulation System., Annu Int Conf IEEE Eng Med Biol Soc. 2020 (2020) 2429–2433. https://doi.org/10.1109/EMBC44109.2020.9176629.

[6] C. Lim, H.S. Ko, S. Cho, I. Ohu, H.E. Wang, R. Griffin, B. Kerrey, J.N. Carlson, Multi-Sensor Feature Integration for Assessment of Endotracheal Intubation, J Med Biol Eng. 40 (2020) 648–654. https://doi.org/10.1007/s40846-020-00541-8.

[7] C.M. Pugh, D.A. Hashimoto, J.R. Korndorffer Jr, The what? How? And who? Of video based assessment, The American Journal of Surgery. 221 (2021) 13–18.

[8] S. McQueen, V. McKinnon, L. VanderBeek, C. McCarthy, R. Sonnadara, Video-based assessment in surgical education: a scoping review, J Surg Educ. 76 (2019) 1645–1654.

[9] E. Yanik, Deep learning for video-based assessment of surgical skills, Rensselaer Polytechnic Institute, 2022.

[10] H. Doughty, D. Damen, W. Mayol-Cuevas, Who's better? who's best? pairwise deep ranking for skill determination, in: Proceedings of the IEEE Conference on Computer Vision and Pattern Recognition, 2018: pp. 6057–6066.

[11] I. Funke, S.T. Mees, J. Weitz, S. Speidel, Video-based surgical skill assessment using 3D convolutional neural networks, Int J Comput Assist Radiol Surg. 14 (2019) 1217–1225. https://doi.org/10.1007/s11548-019-01995-1.

[12] E. Yanik, S. Schwaitzberg, J. Yang, X. Intes, S. De, One-shot domain adaptation in video-based assessment of surgical skills, (2023).

[13] E. Yanik, U. Kruger, X. Intes, R. Rahul, S. De, Video-based formative and summative assessment of surgical tasks using deep learning, Sci Rep. 13 (2023) 1038. https://doi.org/10.1038/s41598-022-26367-9.

[14] G. Lajkó, R. Nagyné Elek, T. Haidegger, Endoscopic Image-Based Skill Assessment in Robot-Assisted Minimally Invasive Surgery., Sensors (Basel). 21 (2021). https://doi.org/10.3390/s21165412.

[15] Y. Ming, Y. Cheng, Y. Jing, L. Liangzhe, Y. Pengcheng, Z. Guang, C. Feng, Surgical skills assessment from robot assisted surgery video data, in: 2021 IEEE International Conference on Power Electronics, Computer Applications (ICPECA), 2021: pp. 392–396. https://doi.org/10.1109/ICPECA51329.2021.9362525.

[16] J.-P. Ainam, K. Qin, G. Liu, G. Luo, View-Invariant and Similarity Learning for Robust Person Re-Identification, IEEE Access. 7 (2019) 185486–185495. https://doi.org/10.1109/ACCESS.2019.2960030.

[17] J. Ngiam, A. Khosla, M. Kim, J. Nam, H. Lee, A.Y. Ng, Multimodal Deep Learning, in: Proceedings of the 28th International Conference on International Conference on Machine Learning, Omnipress, USA, 2011: pp. 689–696.

[18] N. Srivastava, R. Salakhutdinov, Multimodal Learning with Deep Boltzmann Machines, J. Mach. Learn. Res. 15 (2014) 2949–2980.

[19] W. Wang, R. Arora, K. Livescu, J. Bilmes, On Deep Multi-view Representation Learning, in: Proceedings of the 32Nd International Conference on International Conference on Machine Learning - Volume 37, 2015: pp. 1083–1092.

[20] M. Kan, S. Shan, X. Chen, Multi-view Deep Network for Cross-View Classification, in: 2016 IEEE Conference on Computer Vision and Pattern Recognition (CVPR), 2016: pp. 4847–4855. https://doi.org/10.1109/CVPR.2016.524.

[21] H. Su, S. Maji, E. Kalogerakis, E. Learned-Miller, Multi-view Convolutional Neural Networks for 3D Shape Recognition, in: 2015 IEEE International Conference on Computer Vision (ICCV), 2015: pp. 945–953. https://doi.org/10.1109/ICCV.2015.114.

[22] Y.-C. Chen, W.-S. Zheng, J.-H. Lai, P.C. Yuen, An Asymmetric Distance Model for Cross-View Feature Mapping in Person Reidentification, IEEE Transactions on Circuits and Systems for Video Technology. 27 (2017) 1661–1675. https://doi.org/10.1109/TCSVT.2016.2515309.

[23] V.I.J. Strijbis, C.M. de Bloeme, R.W. Jansen, H. Kebiri, H.-G. Nguyen, M.C. de Jong, A.C. Moll, M. Bach-Cuadra, P. de Graaf, M.D. Steenwijk, Multi-view convolutional neural networks for automated ocular structure and tumor segmentation in retinoblastoma, Sci Rep. 11 (2021) 145–190. https://doi.org/10.1038/s41598-021-93905-2.

[24] H. Xu, K. Saenko, Ask, Attend and Answer: Exploring Question-Guided Spatial Attention for Visual Question Answering, in: Computer Vision -- ECCV 2016, Springer International Publishing, Cham, 2016: pp. 451–466.

[25] Z. Yang, X. He, J. Gao, L. Deng, A. Smola, Stacked Attention Networks for Image Question Answering, in: 2016 IEEE Conference on Computer Vision and Pattern Recognition (CVPR), 2016: pp. 21–29. https://doi.org/10.1109/CVPR.2016.10.

[26] J. Lu, J. Yang, D. Batra, D. Parikh, Hierarchical Question-image Co-attention for Visual Question Answering, in: Proceedings of the 30th International Conference on Neural Information Processing Systems, Curran Associates Inc., USA, 2016: pp. 289–297.

[27] A. Vaswani, N. Shazeer, N. Parmar, J. Uszkoreit, L. Jones, A. ~N. Gomez, L. Kaiser, I. Polosukhin, Attention Is All You Need, in: NIPS, 2017.

[28] R.R. Selvaraju, M. Cogswell, A. Das, R. Vedantam, D. Parikh, D. Batra, Grad-CAM: Visual Explanations from Deep Networks via Gradient-Based Localization, in: 2017 IEEE International Conference on Computer Vision (ICCV), 2017: pp. 618–626. https://doi.org/10.1109/ICCV.2017.74.

[29] J.J. Rodriguez, L.F. Higuita-Gutiérrez, E.A. Carrillo Garcia, E. Castaño Betancur, M. Luna Londoño, S. Restrepo Vargas, Meta-Analysis of Failure of Prehospital Endotracheal Intubation in Pediatric Patients, Emerg Med Int. (2020). https://doi.org/10.1155/2020/7012508.

[30] P.E. Sirbaugh, P.E. Pepe, J.E. Shook, K.T. Kimball, M.J. Goldman, M.A. Ward, D.M. Mann, A prospective, population-based study of the demographics, epidemiology, management, and outcome of out-of-hospital pediatric cardiopulmonary arrest, Ann Emerg Med. 33 (1999) 174–184. https://doi.org/10.1016/s0196-0644(99)70391-4.

[31] T. Chen, S. Kornblith, M. Norouzi, G. Hinton, A simple framework for contrastive learning of visual representations, in: International Conference on Machine Learning, 2020: pp. 1597–1607.

[32] J. Johnson, A. Alahi, L. Fei-Fei, Perceptual Losses for Real-Time Style Transfer and Super-Resolution, in: B. Leibe, J. Matas, N. Sebe, M. Welling (Eds.), Computer Vision -- ECCV 2016, Springer International Publishing, Cham, 2016: pp. 694–711.

[33] J. Hu, L. Shen, G. Sun, Squeeze-and-Excitation Networks, in: 2018 IEEE/CVF Conference on Computer Vision and Pattern Recognition, 2018: pp. 7132–7141. https://doi.org/10.1109/CVPR.2018.00745.

[34] A. Paszke, S. Gross, F. Massa, A. Lerer, J. Bradbury, G. Chanan, T. Killeen, Z. Lin, N. Gimelshein, L. Antiga, A. Desmaison, A. Kopf, E. Yang, Z. DeVito, M. Raison, A. Tejani, S. Chilamkurthy, B. Steiner, L. Fang, J. Bai, S. Chintala, PyTorch: An Imperative Style, High-Performance Deep Learning Library, in: Advances in Neural Information Processing Systems 32, Curran Associates, Inc., 2019: pp. 8024–8035.

[35] D.P. Kingma, J. Ba, Adam: {A} Method for Stochastic Optimization, in: 3rd International Conference on Learning Representations, {ICLR} 2015, San Diego, CA, USA, May 7-9, 2015, Conference Track Proceedings, 2015.

[36] A. Hryniowski, X.Y. Wang, A. Wong, Where does trust break down? a quantitative trust analysis of deep neural networks via trust matrix and conditional trust densities, Preprint 2009.14701. (2020).

[37] A. Wong, X.Y. Wang, A. Hryniowski, How much can we really trust you? towards simple, interpretable trust quantification metrics for deep neural networks, Preprint ArXiv:2009.05835. (2020).